\documentclass[article, 10 pt, conference]{ieeeconf}  % Comment this line out if you need a4paper
\IEEEoverridecommandlockouts                              % This command is only needed if 
\UseRawInputEncoding                                    % Needed to meet printer requirements.

\let\labelindent\relax

\usepackage{enumitem}
\setlist{topsep=1pt, partopsep=0pt, parsep=0pt, itemsep=0pt}
\usepackage[
top    = 0.75in,
bottom = 0.75in,
left   = 0.75in,
right  = 0.75in]{geometry}
\usepackage{cite}
\usepackage[skip=10pt plus1pt, indent=35pt]{parskip}
\usepackage{placeins}
\usepackage{textcomp}
\usepackage[dvipsnames]{xcolor}
\definecolor{royalblue}{RGB}{65, 105, 225}
\definecolor{maroon}{RGB}{180, 0, 0}
\definecolor{DarkGreen}{RGB}{0, 100, 0}
\usepackage{booktabs}
\usepackage{multirow}
\usepackage{siunitx}
\usepackage{graphicx}
\usepackage[symbol]{footmisc}
\usepackage{adjustbox}
\usepackage{float}
\usepackage{svg}
\usepackage{array}
\usepackage[T1]{fontenc}
\usepackage{subcaption}
\usepackage{xspace}
\usepackage{scrextend}
\usepackage{caption}
\usepackage{tabularx}
\usepackage{listings}
\usepackage{wrapfig}
\usepackage{calligra}
\usepackage{comment}
\usepackage{soul}
\usepackage{balance}
\usepackage{listings}
\usepackage{xcolor}
\usepackage{caption}
\newcolumntype{A}{ >{\centering\arraybackslash} m{4cm} }
\newcolumntype{B}{ >{\centering\arraybackslash} m{1cm} }
\newcolumntype{C}[1]{>{\centering\let\newline\\\arraybackslash\hspace{0pt}}m{#1}}

\makeatletter
\newcommand\footnoteref[1]{\protected@xdef\@thefnmark{\ref{#1}}\@footnotemark}
\makeatother

\usepackage{amsmath,lipsum} % assumes amsmath package installed
\usepackage{amssymb}  % assumes amsmath package installed
\usepackage{amsfonts}
\usepackage{textgreek}
\usepackage{authblk}
\usepackage{etoolbox}

\makeatletter
\let\NAT@parse\undefined

\let\oldthebibliography\thebibliography
\let\endoldthebibliography\endthebibliography
\renewenvironment{thebibliography}[1]{
  \oldthebibliography{#1}
  \setlength{\itemsep}{-1.75ex plus-.1ex minus-.1ex} % Adjust the value (e.g., -1ex) to change the spacing
}{
  \endoldthebibliography
}

\makeatother

\usepackage[colorlinks=true, citecolor=cyan]{hyperref}  
\def\BibTeX{{\rm B\kern-.05em{\sc i\kern-.025em b}\kern-.08em
    T\kern-.1667em\lower.7ex\hbox{E}\kern-.125emX}}

\title{\LARGE \bf
Anticipate \& Collab: Data-driven Task Anticipation and Knowledge-driven Planning for Human-robot Collaboration
\thanks{*Denotes equal contribution}
}

\author{ Shivam Singh$^{1*}$, Karthik Swaminathan$^{1*}$,  Raghav Arora$^{1*}$, Ramandeep Singh$^{1}$, Ahana Datta$^{1}$, \\ Dipanjan Das$^2$, Snehasis Banerjee$^2$,  Mohan Sridharan$^3$,  Madhava Krishna$^1$
\thanks{$^{1}$ Robotics Research Center, IIIT Hyderabad, India}
\thanks{$^{2}$ TCS Research, Tata Consultancy Services, India}
\thanks{$^{3}$ School of Informatics, University of Edinburgh, UK}
}

\makeatletter
\renewcommand{\@seccntformat}[1]{%
  \protect\csname the#1\endcsname\protect\quad%
}
\makeatother

\renewcommand{\thesection}{\arabic{section}}
\renewcommand{\thesubsection}{\thesection.\arabic{subsection}}
\renewcommand{\thesubsubsection}{\thesubsection.\arabic{subsubsection}}

\begin{document}

\maketitle
\thispagestyle{empty}
\pagestyle{empty}

%%%%%%%%%%%%%%%%%%%%%%%%%%%%%%%%%%%%%%%%%%%%%%%%%%%%%%%%%%%%%%%%%%%%%%%%%%%%%%%%
\begin{abstract}
An agent assisting humans in daily living activities can collaborate more effectively by anticipating upcoming tasks. Data-driven methods represent the state of the art in task anticipation, planning, and related problems, but these methods are resource-hungry and opaque. Our prior work introduced a proof of concept framework that used an LLM to anticipate 3 high-level tasks that served as goals for a classical planning system that computed a sequence of low-level actions for the agent to achieve these goals. This paper describes DaTAPlan, our framework that significantly extends our prior work toward human-robot collaboration. Specifically, DaTAPlan's planner computes actions for an agent and a human to collaboratively and jointly achieve the tasks anticipated by the LLM, and the agent automatically adapts to unexpected changes in human action outcomes and preferences. We evaluate DaTAPlan's capabilities in a realistic simulation environment, demonstrating accurate task anticipation, effective human-robot collaboration, and the ability to adapt to unexpected changes. \\%a substantial reduction ($25.3\%$) in execution time compared with situations without collaboration.  
%\footnote[2]{This project is funded in part by TCS Research India.}
Project website\footnote[4]{This project is supported in part by TCS Research India}: \href{https://dataplan-hrc.github.io}{https://dataplan-hrc.github.io}
% Household agents are capable of performing a wide range of household tasks, either based on user input with constant feedback or autonomously, working towards a common goal of assisting humans. When agents adhere to the latter approach, effective collaboration can be achieved by understanding patterns in daily activities and anticipating the next set of household tasks. While many state-of-the-art methods rely on deep network and sampling-based architectures, these approaches often require extensive amounts of data, limiting their scope. Our framework proposes a novel method that leverages pretrained language models for task anticipation, deterministic classical planners for optimization, and an understanding of the expected distribution of tasks in human-robot collaboration scenarios, with provisions for adaptation in uncertain circumstances. We ground our framework and evaluate its execution within a dynamic simulation environment using CoppeliaSim, demonstrating a maximum reduction of 25.7\% in execution time compared to situations without collaboration. Website Link: \href{https://planology-hrc.github.io}{https://planology-hrc.github.io} 
\end{abstract}
\vspace{-1em}
% \textbf{Color Coding}: \\
% \crule[purple]{10pt}{10pt} : \raghav{Raghav} \\
% \crule[blue]{10pt}{10pt} : \karthik{Karthik}\\
% \crule[olive]{10pt}{10pt} : \shivam{Shivam}\\

\section{Introduction}

\begin{figure}[tb]
  \centering
  \captionsetup{font=scriptsize}
  \includegraphics[width=\columnwidth]{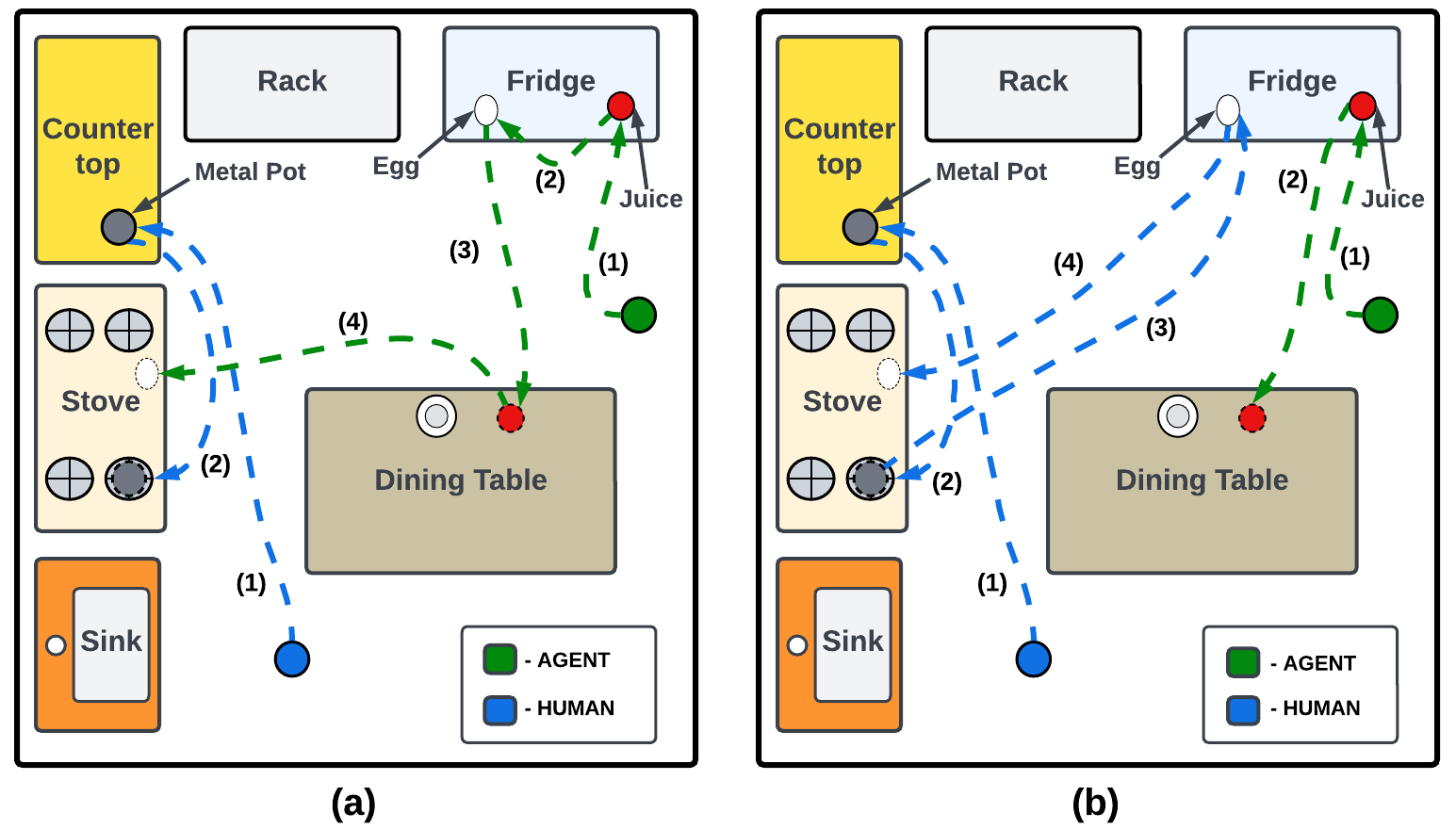}
  \caption{Illustration of "human-robot collaboration with anticipation": (a) agent anticipates (serving task) and collaborates with human, fetching juice from the fridge to the table while fetching the egg that the human cooks in a metal pot and brings to the table; (b) The agent only serves the juice to the table and the human entirely performs the necessary actions needed to cook and serve the egg.}
  \label{fig:hrc_teaser}
  \vspace{-2em}
\end{figure}

% \raghav{Ad-hoc teamwork\cite{StoneKKR10} is defined as the task of collaborating with teammates without pre-coordination. 
% In human-robot collaborative settings, often the robot learns how to act with the full knowledge of the tasks, and the human counterpart. 
% However, in many real-world systems, robots are required to collaborate with humans without prior learning of their working. 
% This work proposes a household robotic agent that can coordinate and work with human collaboration, while adapting to new human agents.}
% \textbf{PARA 1: Idea /Thought about the application of task anticipation + HRC.}\\
Consider a human getting ready to leave from home for work. This involves completing some high-level tasks, e.g., cooking breakfast and serving it at the table in Figure~\ref{fig:hrc_teaser}. Each of these tasks requires the execution of a sequence of actions, e.g., fetch and boil the egg to cook breakfast, and bring the cooked egg and juice to the table. There is a robot (\textit{agent}\footnote[2]{We use the terms "robot" and "agent" interchangeably.}) that can assist in completing these tasks. The tasks can be completed more effectively by anticipating the upcoming tasks, with the agent and the human executing actions to collaboratively and jointly complete all these tasks with minimal effort. This happens in Figure~\ref{fig:hrc_teaser}(a), with the agent (in green) anticipating the serving task, fetching juice from the fridge to the table while fetching the egg that the human (in blue) cooks in a metal pot and (not shown here) brings to the table. There is no such collaboration in Figure~\ref{fig:hrc_teaser}(b), with the agent fetching juice to the table while the human fetches and boils the egg before bringing it to the table. Furthermore, actions may have unexpected outcomes, and changes in human preferences (e.g., the human decides to work from home) may change the tasks to be completed.

%collaborate to jointly  such optimization problems, the agent needs to make sure it parallelizes certain tasks while anticipating the future task, helping it achieve the goal faster. In the realm of computational optimization, this type of "Time and Work" problem poses a fundamental challenge: to determine the time requirements for completing a task when considering the collaborative efforts of the human and the agent operating at varying rates. 

% \textbf{PARA 2: Describe shortly about other used techniques (Writing about HRC framework and implementations as a whole)}\\
\vspace{-1em}
Data-driven methods and models are the state of the art for task anticipation and human-robot collaboration. These methods are resource-hungry, i.e., need considerable computation and training examples, and are opaque, i.e., it is difficult to understand their internal processes. Our recent work provided a basic demonstration of an agent in a household scenario (with no other actors) using a pretrained Large Language Model (LLM) for task anticipation and using classical planning to compute a sequence of finer-granularity actions to jointly achieve these tasks~\cite{anticipate_act}. In this paper, we significantly extend this work to describe a framework (DaTAPlan) that combines data-driven task anticipation and knowledge-driven planning for human-robot collaboration.
%Our work uses a pretrained/fine tuned LLM model to compute a list of high-level tasks anticipated by the agent in a Human-Robot collaboration setting making it more seamless and additionally accommodating uncertainties shown by the human at any given point of execution, making it time inexpensive. Our work leverages the use of classical planners and optimizes on the task distribution between the human and agent at the action level.   \\
% \textbf{PARA 3: Describe about our method and contributions}\\ This data-driven and commonsense reasoning capability of LLM and knowledge-driven planning for a close-to-optimal, collective action representation using classical planners in Human-Robot Collaboration domain have been exploited and our contributions are list below: 
The key characteristics of DaTAPlan are:
\vspace{-1em}
\begin{enumerate}
    \item A pretrained LLM predicts a list of anticipated tasks based on a small number of prompts comprising a partial sequence of tasks in specific user scenarios. 
    %with the capability of adapting to various cases based on the type of user. These anticipated tasks showcase an overall reduction in execution time when human and agent collaborate.
    \item A classical planner reasons with prior knowledge of the action theories of the agent and the human to compute a plan of finer-granularity actions that the agent executes, and the human is expected to execute, to collaboratively achieve the identified list of tasks. %, including  some models of the \textit{jointly by the agent and the human} An intricate and structured PDDL Domain accounting for human and agent actions and effects in the environment.
    \item If the human's action choices, action outcomes, or preferences deviate from expectations as the agent executes its actions, the agent automatically adapts by replanning or generating new task predictions. %A Framework of HRC and Task Anticipation capable of taking into account failure cases.
    \vspace{-1em}
\end{enumerate}
The novelty is in: (a) supporting adaptation to different task patterns with limited prompting to LLMs; (b) computing a plan to jointly achieve the anticipated high-level tasks such that the agent and the human execute actions to collaboratively complete each task; and (c) automatically adapting to unexpected changes in the action outcomes or preferences of human. We use the Planning Domain Definition Language~\cite{pddl} and the Fast Downward solver~\cite{Helmert_2006} for classical planning. We evaluate our framework's capabilities using household scenarios (with multiple tasks, rooms, and objects) in the realistic CoppeliaSim environment. %involving multiple tasks, rooms, type of users, actions for human and robot. 
% We exhibit experiments at all stages of the proposed method and observe reduction of 25\%

\begin{figure*}[h]
  \centering
  \captionsetup{font=scriptsize}
  \includegraphics[width=\textwidth]{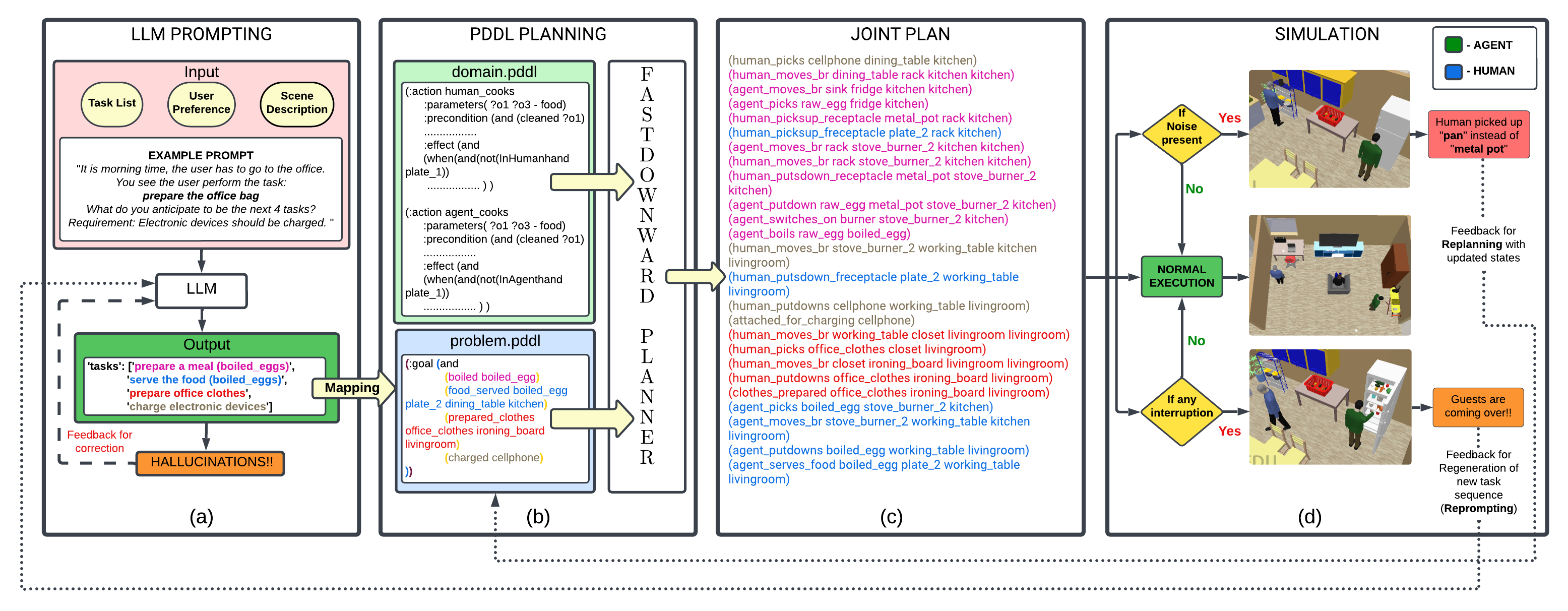}
  \vspace{-1em}
  \caption{Our framework's pipeline: (a) Input prompt contains the list of possible tasks, user preferences, and scene description, along with an example prompt and the corresponding output high-level tasks; (b) High-level tasks predicted by LLM are mapped to PDDL problem description; (c) The FD planner generates a plan of agent's actions and the expected human actions; (d) Deviations of the human from the expected plan are noted and used to trigger replanning when appropriate.}
  \label{fig:pipeline}
  \vspace{-1em}
\end{figure*}

\section{Related Work}
% \cite{Suriadinata}
% \cite{pddl}
% \cite{PDDLbook}
% \raghav{Should we write about LLMs as planners? (Since we don't have time to make comparisons now)}
% \subsection{LLMs as planners}
% \cite{code-as-policies}
% \cite{progprompt}
% \cite{valmeekam2023planbench}
% \cite{TidyBot}
% \cite{zhao2023large}
% \cite{llm-zero}
% \cite{birr2024autogptp}
% \raghav{Almost the same as ICRA paper!}

%\textbf{Human-Robot Collaboration:}
\vspace{-0.5em}
State of the art research in \textit{human-robot collaboration} focuses on teleoperation, shared autonomy, or collaboration~\cite{smith2022,lee2023,kurt2022}. There have been promising breakthroughs in perception, learning, task planning, and adaptive control~\cite{garcia2022,park2024,wang2022,kim2021}, and work on improving collaborative interactions that enhance efficiency and quality of life~\cite{chen2024
% ,patel2023
,li2022,yang2021,liu2024}. Different data-driven formulations represent the state of the art for human-robot collaboration and task anticipation~\cite{MRLC,9982251}. However, safety, trust, adaptability, real-time perception, and integration of human feedback remain open problems~\cite{brown2021}. %along with real-time perception and integration of human feedback~\cite{zhang2024}. Prior work has concentrated on improving collaborative interactions between humans and household robots to enhance efficiency and quality of life \cite{johnson2023,chen2024,patel2023,li2022,yang2021,liu2024}. 
% This has led to the development of frameworks and systems enabling robots to assist with various household tasks, including cleaning, cooking, organizing, and scheduling \cite{brown2021}. These efforts drive the evolution of HRC technologies, providing innovative solutions to enhance robots' effectiveness and autonomy as collaborators in domestic settings \cite{chen2023,patel2021}.
%\textbf{LLM+PDDL Architecture:}

\vspace{-0.5em}
Planning Domain Definition Language (PDDL) has been widely used to encode prior knowledge for planning problems~\cite{pddl}. Within automated task planning methods, the process of defining the planning problem and prior domain knowledge (i.e., domain models) is labor-intensive and relies on closed-world assumptions, limiting adaptability to dynamic environments. There has been considerable research in learning the domain models for planning, with more recent work using LLMs for this purpose~\cite{valmeekam2022large}. LLMs have also been used to produce goal states achievable by classical PDDL-based planners \cite{liu2023llmp, xie2023translating}. and for generating diverse plans or translating natural language to structured planning problems~\cite{joublin2023copal, plancollabnl2024}. However, integrating learning methods with planning systems while ensuring that sound and executable plans are produced, remains challenging~\cite{valmeekam2023planbench, hirsch2024whats}. Recent studies have also used LLMs for task planning in complex domains~\cite{silver2023generalized, birr2024autogptp, silver2022pddl, Lin:NIPS23}, including scene rearrangement~\cite{zhao2023large}, but there is a growing body of research to show that LLMs are not really appropriate for planning in the classical sense~\cite{valmeekam2022large}.

\vspace{-0.5em}
There is well-established research in monitoring action outcomes and adapting to unexpected outcomes~\cite{ghallab:plan04}. Recent work in human-robot collaboration uses human behavior models to detect unexpected behavior and has the agent replan to prevent failure~\cite{9812236}.
%Some literature incorporates error recognition and correction in execution systems. 
Even systems that use LLMs for planning include a method for detecting collisions or hardware failures~\cite{joublin2023copal}, or use probabilistic sequential decision-making for monitoring~\cite{Yang2021AHP}.
Our framework enables the agent to adapt if the human's action choices, action outcomes, or preferences deviate from expectations.% from the computed plans in the form of unexpected  plan execution deviations: (i) Human interruptions prompt reprompting for new goals, (ii) Action noise triggers feedback and replanning.

\vspace{-0.5em}
Our prior work enabled an agent in a household scenario to use a pre-trained LLM for task anticipation, with classical planning used to compute and execute a sequence of finer-granularity actions to jointly achieve these tasks~\cite{anticipate_act}. Here we extend this approach to human-robot collaboration, with planning directing the agent and the human to collaboratively execute actions to complete each task and enabling the agent to adapt when the human deviates from the plan.

% Deviating from existing literature, in this paper, we extend the previous work \cite{anticipate_act} by employing LLMs to anticipate customized household tasks and facilitate human-robot collaboration. We evaluate LLM's adaptability for changing task requirements and introduce a closed-loop feedback framework which will trigger replanning when the robot detects deviations from expected tasks. This proactive approach ensures smooth collaboration, rectifying unintended actions and achieving shared objectives efficiently.
% Building upon our previous research \cite{anticipate_act}, in this paper, we leverage LLMs to anticipate tasks tailored to specific households and user preferences, integrating human-robot collaboration into daily activities. Additionally, we assess the adaptability of LLMs in response to unforeseen changes in task requirements. We have developed a closed-loop feedback framework that triggers replanning whenever the robot detects deviations from its anticipated actions within the environment. This proactive approach allows for the rectification of unintended actions, thereby establishing a robust framework to facilitate smooth human-robot collaboration in achieving shared objectives within household tasks.
\section{Problem Formulation and Framework}
\label{sec:framework}
% \raghav{
% \begin{itemize}
%     \item Explain the environment (number of rooms, objects, receptacles, types of objects, etc)
%     \item Describe the task list and action space
%     \item Explain the costs assigned to actions (distribution between human and agent costs)
% \end{itemize}
% }
Consider a household with two actors, \textit{human} ($\mathcal{H}$) and \textit{agent/robot} ($\mathcal{R}$). The objective is to complete a routine of high-level tasks $\mathcal{Q} = \{\tau_1, \tau_2,..., \tau_n\}$ although the entire routine is not known in advance and can change over time. Completion of each $\tau_i \in \mathcal{T}$, a known list of high-level tasks such as \textit{cook breakfast} and \textit{do the laundry}, requires a sequence of finer-granularity actions $\{a_1$, $a_2$, ...., $a_k\}$ to be executed, e.g., to \textit{cook breakfast}, it is necessary to \textit{go to the fridge}, \textit{fetch egg to the stove}, and \textit{boil the egg}. Since $\mathcal{Q}$ can change over time, an actor usually tries to complete one task at a time at minimum cost (or time, effort). However, the tasks can be completed more efficiently if $\mathcal{R}$ and $\mathcal{H}$ anticipate upcoming tasks and collaborate to complete them. The agent has an action theory, $\mathcal{M}_\mathcal{R}$, describing preconditions and effects of its actions, and a similar theory $\mathcal{M}_\mathcal{H}$ describing its expectations of human behavior. % for the household to construct plans and achieve the goal state. $M_\mathcal{R}$ and $M_\mathcal{H}$ contain set of \textit{grounded} agent and human actions respectively. The description of these actions are defined in PDDL domain description.
% We use a fast-downward model which based on PDDL \cite{pddl}, a general planning domain definition language. 
% However, because of the presence of the human in the same environment, the robot needs to generate collaborative plans; it creates a plan for itself ($M_\mathcal{R}$) along with an expected plan for the human ($M_\mathcal{H}$). 
There is no explicit communication between $\mathcal{R}$ and $\mathcal{H}$ and we only control $\mathcal{R}$'s action choices. Actions can be non-deterministic but the domain state is assumed to be fully observable.

% For generating $M_\mathcal{R}$ and $M_\mathcal{H}$, we utilize LAMA\cite{lama} alias of the Fast Downward \cite{Helmert2006} planning system.
\vspace{-0.5em}
Figure \ref{fig:pipeline} outlines our framework, DaTAPlan, which combines data-driven task anticipation and knowledge-driven planning (with monitoring) for human-robot collaboration. In Figure~\ref{fig:pipeline} (a), an LLM is prompted with a partial task sequence to obtain a sequence of anticipated tasks. These tasks become goals in a PDDL problem file in Figure~\ref{fig:pipeline}(b). A classical planner uses this problem description and a domain description to compute a joint plan of low-level actions for the agent and expected actions for the human---Figure~\ref{fig:pipeline}(c). The execution of these actions is simulated in CoppeliaSim~\cite{coppeliasim} in Figure~\ref{fig:pipeline}(d). The agent monitors any deviation from the plan and adapts accordingly. We describe the individual components of DaTAPlan below.

\subsection{LLM-based Task Anticipation}
\label{sec:framework-llm}
\begin{figure}[tb]
  \centering
  \includegraphics[width=\columnwidth]{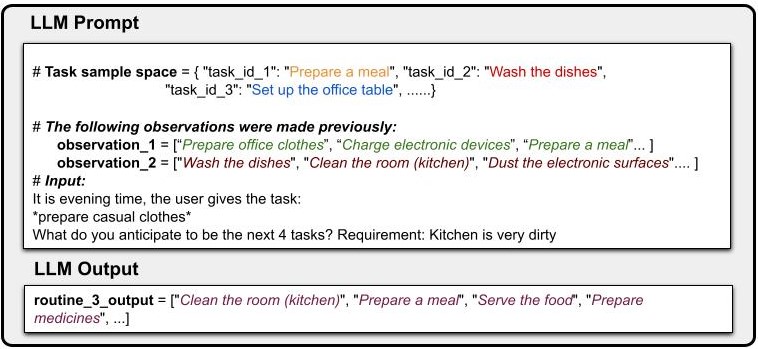}
  \caption{Few shot prompting with LLMs.}
  \label{fig:llm_prompt}
    \vspace{-1em}
\end{figure}

%In our setting, an intelligent agent receives a high-level task, $\tau_i$ ($\in$ $\mathcal{T}$) from a user. The agent is capable of executing only the set of tasks belonging to it's task-space $\mathcal{T}$~\footnote[4]{\label{fn: website}You can find the list of tasks, in-context examples and LLM outputs on our website.}. The tasks from $\mathcal{T}$ are mapped to goal description of PDDL problem file. 
A pretrained LLM is tuned with two types of prompts: (i) \textit{few shot}; and \textit{chain of thought}. In both cases, the inputs include $\mathcal{T}$ and a JSON scene description. In the former, the input also includes 2-3 prior observations of user task patterns---see Figure~\ref{fig:llm_prompt};, whereas the latter considers two in-context examples~\cite{brown_few_shot} with step-by-step guidance~\cite{wei2023chainofthought} to understand user patterns. In both cases, the output is a sequence of anticipated tasks, with hallucinated tasks outside $\mathcal{T}$ not used for planning. Please see our project website for more examples. We show later (Section~\ref{sec:expres-results}) that the LLM's task predictions match human expectations.

%We observe that language models match human expectations at identifying patterns when asked to think step-by-step, with a few contextual example. While the output of LLM is restricted to a known set $\mathcal{T}$, it takes the open-world description of the environment, with the state at a given time, and uses the knowledge to accurately anticipate future tasks. 
% We observe that providing the dataset $\mathcal{T}$ to the LLM along with a feedback mechanism ensures its output belong to the closed set. 
\vspace{-1em}
Apart from anticipating future tasks, the LLM  also receives user feedback (during execution) if there is a change in tasks from a usual pattern, e.g., if the user shares that \textit{guests are coming over}, then LLM is re-prompted to retrieve a fresh set of anticipated tasks. %Similar to the previous example, we measure the language models at anticipating tasks based on a specific re-prompting scenario against human expectations for the same statement.

\subsection{Task and Motion Planning}
\label{sec:framework-tamp}
The sequence of anticipated high-level tasks $\mathcal{Q}_A = \{\tau_1, \tau_2,...\tau_n\}$ from the LLM is mapped to goal state $\mathcal{G}$ in the planning framework---see Figure~\ref{fig:pipeline}(b). The next component of the framework computes a sequence of finer-granularity actions to be executed by $\mathcal{R}$ and $\mathcal{H}$ to achieve $\mathcal{G}$.

%Each task $\tau_i$ contains a set of actions $a_m^{\mathcal{R}}$, $a_n^{\mathcal{H}}$ where $a_m^{\mathcal{R}}$ is the $m^{th}$ action taken by the agent, and $a_n^{\mathcal{H}}$ is the $n^{th}$ action taken by the human.

% For instance, in the Fig \ref{fig:pipeline}, the task \textit{prepare a meal} involves actions like {\small \texttt{(human\_putsdown\_receptacle metal\_pot stove\_burner\_2 kitchen)}} , where the human puts down the metal pot on stove; and {\small \texttt{(agent\_putdown raw\_egg metal\_pot stove\_burner\_2 kitchen)}} where the agent puts down egg in the metal pot on the stove. 
% For the first part of our experiments, we assume that the human and agent will collaborate and follow their individual actions. Then, in section~\ref{sec: h4} we show a deviation in two scenarios: (a) Deviation from the anticipated routines in case of interruption from the user; (b) Deviations in actions executed by the actors.
% We show (b) in \ref{sec: h4} for specifically the `pick-up' actions of the human.
% We encode the environment description in PDDL \cite{pddl} inspired by~\cite{anticipate_act}. 
% This domain mainly contains the description of human and agent actions describing \textit{preconditions} and \textit{effects} of these actions. 
% We qualitatively demonstrate the execution of tasks, and motion planning in the CoppeliaSim simulator.\\

\noindent
\textbf{Task planning.} To generate a sequence of actions to achieve $\mathcal{G}$, the \texttt{domain} and \texttt{problem} are created in PDDL. 

\noindent
The domain description $\mathcal{D} = \langle\mathcal{D}_\mathcal{R}, \mathcal{D}_\mathcal{H}\rangle$ comprises the domain descriptions of the agent ($\mathcal{D}_\mathcal{R}$) and the human ($\mathcal{D}_\mathcal{H}$). $\mathcal{D}_\mathcal{R} = \langle \mathcal{S}_\mathcal{R}, \mathcal{M}_\mathcal{R} \rangle$ comprises signature $\mathcal{S}$ and action theory $\mathcal{M}_\mathcal{R}$.
$\mathcal{S}$ includes \textit{types}, \textit{constants}, and \textit{predicates}. \textit{Predicates} include \textit{fluents}, which can change over time due to \textit{actions}; and \textit{statics}, which remain unchanged. For instance, in Figure~\ref{fig:pipeline}(c), actions such as \textit{human\_picks} and \textit{agent\_picks} modify fluent \textit{obj-at} that determines the location of objects. $\mathcal{M}_\mathcal{R}$ specifies each action (of the agent) in terms of its parameters, preconditions for the action to be executed, effects that will be true once the action is executed, and the action's cost. In a similar manner, $\mathcal{D}_\mathcal{H} = \langle \mathcal{S}_\mathcal{H}, \mathcal{M}_\mathcal{H} \rangle$, is the agent's estimate of the human's domain description. Here, we assume that these descriptions are accurate; the revision of this description is left to future work. 
%e.g., action \textit{human\_picks cellphone} has the precondition that the human and cellphone be in close proximity. 

\vspace{-1em}
The \texttt{problem} $\mathcal{P} = \langle\mathcal{O}, \mathcal{I}, \mathcal{G} \rangle$ describes a specific scenario in terms of a set of specific objects ($\mathcal{O}$), the initial state $\mathcal{I}$ comprising ground literals of the \textit{fluents} and \textit{statics}, and a goal description $\mathcal{G}$ in the form of relevant ground literals. To create suitable scenarios,  we designed a complex household domain with a range of actions and objects (72 predicates and 88 actions); \textit{this is more complex than commonly used planning benchmarks and our prior work}. $\mathcal{I}$ is estimated from sensor inputs and $\mathcal{G}$ includes the anticipated tasks. Example problem and domain descriptions are available in our open-source project website.

The planning task is to compute a sequence of actions $\pi = \langle a_1, \ldots, a_K \rangle$ that takes the system from $\mathcal{I}$ to a state where $\mathcal{G}$ is satisfied. Some actions in this sequence are to be executed by the agent, while the human is expected to execute the other actions.
% Each task $\tau_i$ has some preconditions $\{p^{i}_{k}\}$ where $ k \in \mathbb{R}$, and the task is deemed completed after each precondition is satisfied.
To compute the plan, $\mathcal{D}$ and $\mathcal{P}$ are given to the planner that tries to minimize the total cost of actions required to achieve $\mathcal{G}$. We use the LAMA\cite{lama} alias of the Fast Downward system~\cite{Helmert2006} to compute the plan. To reduce costs, we opt for \textit{satisficing} instead of \textit{optimal} configurations of the planner. To find the \textit{satisficing} plan $\pi^*$ that minimizes cost, we define the objective function as:
\begin{align}
    \label{eqn:plan-cost}
    \pi^* = \arg \min_{\pi} C(\pi), \quad C(\pi) = \left( \sum_{m=0}^M c_{m}^\mathcal{R} + \sum_{n=0}^N c_{n}^\mathcal{H}\right)
\end{align}
where $c_m^\mathcal{R}$ is the cost of the agent's action $a_m^\mathcal{R}$ in plan $\pi$, and $c_n^\mathcal{H}$ is the cost of the human's action $a_n^\mathcal{H}$ in plan $\pi$. The cost of each action corresponds to the time taken to execute it. The optimal plan $\pi^*$ minimizes $C(\pi)$ for both actors.

\noindent
\textbf{Human-robot collaboration.}
% The final satisficing plan obtained $\pi*$ = ($a_1$, $a_2$, \..., $a_k$ ) contains actions for both the human and the agent, involving collaboration and parallelization over multiple actions for the common goal.
% We can consider the example shown in the figure \ref{fig:pipeline}, where we see human-robot collaboration in the tasks \textit{prepare a meal}, and \textit{serve the food}.
% As per the plan generated, the human moves the \textit{metal-pot} to the \textit{stove-burner}, while the agent fetches \textit{eggs} from the \textit{fridge} to boil and serve them on the \textit{dining-table}. Because of the mutual collaboration of actions, we see an optimized plan where the join cost of execution is minimized.
The computed plan $\pi^{*} = \langle a_1$, $a_2$, ..., $a_k\rangle$ represents a collaboration between the human and the agent. 
%Based on its model of the human's action theory, the agent expects the human to execute some of the actions. 
For example, the goal states in Figure~\ref{fig:pipeline} involve \textit{preparing and serving breakfast of boiled eggs}, \textit{preparing office clothes}, and \textit{charging the cellphone}. To prepare breakfast, while the agent fetches eggs from the fridge nearby to the stove, the human is expected to bring a metal pot to the stove. Doing so will result in the task being completed in the least amount of time; Figure~\ref{fig:hrc_teaser}(a) is an illustration of this "collaboration" setting. There is no explicit communication between the human and the agent. While the agent will execute the relevant actions in the plan, there is no guarantee that the human will execute the assigned actions. Figure~\ref{fig:hrc_teaser}(b) is an illustration of the "no-collaboration" setting in which the agent and the human are assigned specific actions that do not minimize the overall cost. We experimentally compare the collaboration and no-collaboration settings in Section~\ref{sec:expres-results}.

%This demonstrates the effective collaboration between the human and the robot, ensuring that the time taken to complete the goal state is minimized. In a no collaborative scenario, they work independently to achieve the tasks present in the goal state. 
%A comparison of collaboration vs no collaboration is given in . \\
%As discussed previously, in collaborative setting, the human's actions may deviate from the expected plan $M_\mathcal{H}$. In this case, our framework recovers by re-planning and updating the expected human actions in $M_\mathcal{H}$ based on the observed deviations. \\

\noindent
\textbf{Adaptation to unexpected situations.}
DaTAPlan includes an approach that enables the agent to adapt if the human's action choices, action outcomes or preferences deviate from the expectations. Recall that the plan $\pi^{*}_{i}$ will succeed only if both the agent and the human execute the actions assigned to them. The agent will meet this requirement, but we cannot guarantee that the human will do so. In addition, the human's task-level preferences may change and make the anticipated tasks irrelevant. For example, consider an agent executing actions to achieve the anticipated tasks of preparing and serving breakfast, and preparing the human's office clothes. This agent may find that the human was unable to bring the metal pot to the stove (to cook breakfast), requiring the agent to generate a new plan, or may be told that the human is no longer going to the office, requiring the LLM to provide new predictions of anticipated tasks.

\vspace{-1em}
In this paper, we illustrate our adaptation approach in the context of a specific kind of unexpected outcome: when the human does not execute a planned \textit{pick up} action or the execution of this action does not result in the desired object being picked up. For instance, the plan shown in Figure~\ref{fig:pipeline} involves boiling eggs in a metal pot. Here, the agent expects a human to bring the metal pot to the stove (see Figure~\ref{fig:Action}) while it fetches the eggs from the fridge to the stove. If the human does not pick up the metal pot, the subsequent steps of the plan will not achieve the part of the goal related to cooking breakfast. Although the agent and the human do not communicate with each other, the agent believes that it has a good model of the human's domain description, and the system provides full observability. The agent thus adapts by replanning from the current state. We experimentally compare performance with and without this adaptation strategy in Section~\ref{sec:expres-results}. 

\vspace{-1em}
We also illustrate the ability to adapt to unforeseen changes in the human's preferences. In this case, the agent directly prompts the LLM for a new sequence of anticipated tasks before replanning to achieve the revised goal.

\begin{figure}[tb] % Use figure* to span both columns
\captionsetup{font=scriptsize}
\begin{center}
    \begin{minipage}{0.45\textwidth} % Adjust the width as needed
        \begin{lstlisting}
(:action agent_boils
 :parameters (?o - toboil)
 :precondition(and
              @(item_in ?o metal_pot stove kitchen)@
              (agent_near stove kitchen)
              (agent_switched_on burner stove kitchen)
              (not(boiled ?o)))
 :effect(boiled ?o))
        \end{lstlisting}
    \end{minipage}
    \setlength{\abovecaptionskip}{-3pt}
    \setlength{\belowcaptionskip}{0pt}
    \caption{Action for \textit{boiling} an item ?o\protect\footnotemark. Precondition: item must be in the metal pot.}
    \label{fig:Action}
    \vspace{-1.5em}
\end{center}
\end{figure}
\footnotetext{Due to space constraints, not all preconditions and effects are shown.}

\noindent
\textbf{Motion Planning.}
The low-level actions from the Task Planning are interpreted and are provided to the actors through CoppeliaSim Remote API. For actions that entail movement between locations, the nearest free space goal position is given to the OMPL BiTRRT Planner~\cite{sucan2012the-open-motion-planning-library} with 10000 maximum search point nodes and 0.1 seconds for search duration for each simulation pass as search parameters. For actions such as \textit{picking} and \textit{cooking}, the object's goal position is used to control the movement of the arm using the Inverse Kinematics method provided by the CoppeliaSim IK Plugin.  
%%%%%%%%%%%%%%%%%%%%%%%%%%%%%%%%%%%%%%%%%%%%%%%%%%%%%%%%%%%%%%%%%%%%%%%%%%%%%%%%%%%%%%%% 
\section{Experimental Sections and Results}
%%%%%%%%%%%%%%%%%%%%%%%%%%%%% EXPERIMENTAL SECTION %%%%%%%%%%%%%%%%%%%%%%%%%%%%%%%%%%%%%
\noindent
We experimentally evaluated four hypotheses related to the performance of DaTAPlan:
\vspace{-0.75em}
\begin{itemize}
\item[\textbf{H1:}] LLMs can accurately anticipate future tasks based on a small number of contextual examples.%(*rephrase*)
\item[\textbf{H2:}] Combining task anticipation and action planning substantially improves efficiency of planning and execution in human-robot collaboration scenarios compared with using just the classical planner. % based plans can be used by multi-agent settings leading to human-robot collaboration (*rephrase*).
\item[\textbf{H3:}] Human-robot collaboration results in more efficient goal attainment compared with no active collaboration.
  % Different scenarios: speed of human vis-a-vis speed of agent. Hypotheses 2 and 3 will work with varying speeds of agents. \commentm{This is more of a parameter setting to be explored in experiments and not necessarily a hypothesis to be tested...}
\item[\textbf{H4:}] Agent is able to automatically adapt to unexpected changes in action outcomes and preferences of humans. 
  % \item[\textbf{H5}] Our framework performs better for anticipation and planning than LLM based anticipation and planning.  
  % \item[\textbf{H5}] Our framework supports incremental learning of the action costs based on interactions with the environment. %can take feedbacks and learn the environments updating the costs on the fly.
  \vspace{-1.5em}
\end{itemize}
The Fast-Downward system \cite{Helmert_2006} provides different configurations: \textit{lama}, \textit{seq-sat-fdss-2018}, and \textit{seq-sat-fd-autotune-1}. We experimentally determined that \textit{lama} provides the best performance and used it for all experiments reported here.

% \vspace{-0.5em}
%%%%%%%%%%%%%%%%%%%%%%%%%%%%%%%%%%%%%%%%%%%%%%%%%%%%%%%%%%%%%%%%%%%%%%%%%%%%%%%%%%%%%%%%% 
\subsection{Experimental Setup}
\label{sec:expres-setup}
%%%%%%%%%%%%%%%%%%%%%%%%%%% EXPERIMENTAL SETUP %%%%%%%%%%%%%%%%%%%%%%%%%%%%%%%%%%%%%%%%%%
\noindent
We begin by describing the experimental set up process.

\noindent
\textbf{LLM Prompting.}
% Different large language models like `gemini-1.0-pro-latest` \cite{geminiteam2023gemini} by Google, `gpt-3.5` \cite{gpt} by OpenAI and `claude-3-opus` by Anthropic are used for evaluating Hypothesis \textbf{H1}. 
We evaluated H1 quantitatively using different LLMs: Gemini Pro~\cite{geminiteam2023gemini}, Claude 3~\cite{anthropic2023claude3} and GPT-4~\cite{OpenAI2023GPT4TR}. We also extended the diversity of household tasks introduced in our prior work~\cite{anticipate_act}, to obtain 16 global tasks. %$\tau_i$ like `prepare a meal', and `wash the dishes'. 
To explore specific user task patterns, we created two households, each with five different scenarios (e.g., tasks related to different time of day), in which the sequence of task execution is different. \textit{Household-1} is characterized by orderly task execution, whereas in \textit{Household-2} immediate needs are prioritized with tasks performed as they come. 

\vspace{-1em}
We collected and used actual human expectations as the \textit{ground truth}. 
%The anticipation outputs from different LLMs were compared with that of human expectations. 
%For this, we create a set of five different scenarios for each household. 
% These scenarios capture diverse situations that might influence task selection (e.g., taking medicines after dinner, cleaning up the house for guests).
Specifically, for each scenario, human expectations were recorded from 11 humans, with each of them asked to anticipate four tasks at a time. The humans were provided the same information as the LLMs. %with a simplified information about the task patterns. 
There is variability in the responses of different humans to the same scenario, and it is often hard to identify the "correct" response. Hence, we obtained the ground truth for each scenario by selecting four tasks with the highest frequency from all human responses for that scenario. 

\vspace{-1em}
Since computing the frequency of tasks may lead to the loss of information about the sequencing of tasks, we (instead) measured overlap.
% Each LLM was prompted $25$ times for each scenario and the outputs were compared with the human expectations using the following metrics:
%\subsubsection*{LLM Metrics}
Let $g_i$ be the set of (most frequent) tasks in the human responses (ground truth) for the $i^{th}$ scenario, and $l_i$ be the sets of tasks predicted by the LLM.; $g_i \cap l_i$ denoted the set of tasks that appear in both $g_i$ and $l_i$. We then used two measures to evaluate \textbf{H1}.
\vspace{-0.7em}
\begin{itemize}
\item \textit{Mean Overlap:} the average overlap between the ground truth and the LLM responses. 
% It is calculated by summing the overlap values for each pair of LLM response and ground-truth anticipation and then dividing by the total number of response pairs.
  \vspace{-0.2em}
  \begin{align}
  \label{eqn:mean-overlap}
  \text{Mean Overlap} = \frac{\sum_{i=1}^{n} |g_i \cap l_i|}{n}
  \end{align}
    where $| \cdot |$ denotes the cardinality of a set, and $n$ is the total number of response pairs.
  \vspace{-0.1em}
\item \textit{$\geq50\%$ and $\geq75\%$ Overlap:} the proportion of LLM outputs that have an overlap of at least $50\%$ or $75\%$ with the ground truth. This is calculated as:
  \vspace{-0.3em}
  \begin{align}
  \label{eqn:overlap}
    \geq \text{Overlap} = \frac{\sum_{i=1}^{n} I(|g_i \cap l_i| \geq k)}{n}
      % \label{eq: overlap50}
  \end{align}
where $k \in \{2, 3\}$, and $I(\cdot)$ is the indicator function defined as: $I(x) = 1$ if $x$ is true, and $0$ if $x$ is false. Note that $50\%$ overlap implies two tasks in common between LLM response and the ground truth, and $75\%$ overlap implies three common tasks.
\end{itemize}
% \[
% I(x) = \begin{cases}
%     1 & \text{if } x \text{ is true} \\
%     0 & \text{if } x \text{ is false}
%   \end{cases}
% \]
In these measures, $n$ is the number of response pairs, i.e., number of scenarios multiplied by the number of LLM prompts. We collected data for five scenarios from 11 humans, and prompted each LLM 25 times: $n = 5\times 25 = 125$.
% Since we are anticipating for 4 tasks at a time, $\geq 50 \%$ overlap means at least 2 tasks common between LLM and human responses, and $\geq 75 \% $ overlap means at least 3 tasks common respectively.\\

% \karthik{
% \begin{itemize}
% \item \st{Explaning the structure of prompts}( Resource Availability and \st{User preferenced based prompts}) 
% \item \st{Mentioning about the metrics ( LLMs vs Human GT outcome, order?, Recovery from resource availability setup)} 
% \item \st{In the case of reprompting}
% \end{itemize}
% }

\noindent
\textbf{Planning.}
The household environment in which the agent and human operated had four rooms: \textit{Bathroom}, \textit{Kitchen}, \textit{Storeroom}, and \textit{Livingroom}. As stated in Section~\ref{sec:framework-tamp}, the mapping of the properties of this environment (and the agent, human) in PDDL had 72 predicates and 88 actions, with 39 \textit{human-specific} actions, 39 \textit{robot-specific} actions, and 10 actions common to both. There were 17 \textit{types} of objects.

\vspace{-1em}
The default domain description had the agent and the human collaborating to achieve the desired goals. To mimic lack of collaboration between the human and the agent, plans were generated separately for the human and the agent with the corresponding domain description only containing the human-specific and the agent-specific actions (respectively).
%We have a common PDDL domain with action descriptions of both the human and the agent, this domain is used for generating collaboration plans. For no-collaboration scenarios, we use our \textit{human-specific} and \textit{agent-specific} domains. We compare the collaborative and non-collaborative scenarios plans in section \ref{sec: h3}.\\

\vspace{-1em}
Plans were computed to minimize the total cost. We determined action costs based on four factors: (i) \textit{distance to the target location}, encoding the principle that the actor closest to a location should move to it; (ii) \textit{object type}, encoding the preference that humans handle fragile objects if possible because it is harder for the agent to safely grasp and move such objects; (iii) \textit{task completion time}, encoding prior knowledge that humans are more efficient with some complex tasks such as cooking while the agent is more efficient with some tasks such as cleaning; and (iv) \textit{action priorities}, encoding the principle that an agent should handle the more repetitive tasks such as fetching objects. Note that the planner often had to trade-off between these factors when computing a plan, e.g., the agent is closer than the human to a fragile object that needs to be moved to a different room.
\vspace{-1em}
Next, we used two measures to evaluate the performance of the planning framework.
\vspace{-0.9em}
\begin{itemize}
\vspace{-0.3em}
\item \textit{Execution cost:} the execution cost of plan $\pi$ is defined as the sum of the costs of all actions executed by the agent ($\mathcal{R}$) and the human ($\mathcal{H}$) while following $\pi$; this is computed as shown in Equation~\ref{eqn:plan-cost}.
% \vspace{-0.3em}
% \setlength{\abovedisplayskip}{0pt}
% \setlength{\belowdisplayskip}{0pt}
% \setlength{\abovedisplayshortskip}{0pt}
% \setlength{\belowdisplayshortskip}{0pt}
% \begin{align*}
% \,C(\pi^*) = \sum_{m=0}^M c_m^\mathcal{R} + \sum_{n=0}^N c_n^\mathcal{H}\,
% \end{align*}
% % \vspace{-0.3em}
% where $c_m^{\mathcal{R}}$ is the cost of the $m^{th}$ action executed by the agent in the plan $\pi^*$, $c_n^{\mathcal{H}}$ is the cost of the $n^{th}$ action executed by the human in the plan $\pi^*$, $M$ is the total number of actions executed by the agent, and $N$ is the total number of actions executed by the human.
% The overall cost translates to the time (in seconds) it would take for both the agent and the human to complete the execution of tasks in the home.
\item \textit{Plan length:} the length of a plan is the total number of actions in the plan, computed as the sum of the number of actions executed by the agent and the human.
%     $|\pi^*| = M + N$; 
% % \begin{align*}
% %     |\pi^*| = M + N
% % \end{align*}
% where $M$ and $N$ are as defined above.\\
\end{itemize}

  % \item In the case of replanning (The planner creates a joint plan involving actions of both human and agent. This plan is considered to be an observation by the robot or formally an expectation of the robot in completing a goal state.)

% The PDDL domain, alongside the problem file containing specific goals, is provided to the planner. The planner generates a collaborative plan consisting of human and robot actions. 
% The human actions included in this plan represent what the robot anticipates the human will do. 
% It's important to note that the plan is derived from the initial state of the environment \textemdash as perceived by the agent\textemdash and the goal state.
%\vspace{-0.5em}
\noindent
\textbf{Adaptation.}
As with collaboration, the standard operation of DaTAPlan had the agent adapting to unexpected changes in the human's action outcomes and preferences as described in Section~\ref{sec:framework-tamp}. Specifically, any unexpected outcome of a \textit{pick up} action executed by a human resulted in a new plan being generated from the current state. The agent then executed the actions (in the plan) allocated to it and expected the human to do the same. The planning and execution costs of this new plan were added to the plan(s) computed and executed so far. In a similar manner, a change in the human's high-level preference resulted in a new sequence of anticipated tasks being generated by the LLM, followed by a new plan. 

%In Section \ref{adanoise}, we outline how our system adapts to unexpected situations and the consequences when there is no adaptation to noise. The initial noise-free plan $\pi^{*}_{i}$ generated by the agent contains multiple human actions that are prone to noise. If the human performs an action that is not expected by the agent, replanning occurs; the agent extracts all state changes up to the point of noise occurrence, then updates these changes in the PDDL problem description. These changes include updated object locations, the state of human etc. It then signals the human about the potential mistake, prompting a possible correction. After updating the state, the agent generates a new collaborative plan $\pi^{*}_{a}$ to achieve the goal states initially defined. In our experiments, we observed that the execution cost after adaptation, $C(\pi^{*}_{a})$, is higher than the initial cost estimate, $C(\pi^{*}_{i})$, due to the additional actions required to correct for the noise.\\
\vspace{-1em}
We simulated the lack of such an adaptive approach by having the agent and the human continue executing the actions allocated to them (to the extent possible) even when the human's action did not have the expected outcome. The agent became aware of the unexpected outcome only when the current plan resulted in failure, at which point a new plan could be generated. In  this case, the time made available to the agent and the human to compute and execute plans matched the time taken to achieve the goal when the adaptation approach was used. The performance measure was the fraction of the goal achieved in the time available.

%In a non-adaptive case, the agent remains unaware of the human's mistakes, resulting in a failure to reach the goal states. Due to the lack of feedback between the human and the agent, the agent will remain unaware even if the human distracts or deviates from the expected path multiple times, ultimately leading to a decline in the success rates of the goal plan. These results are shown in the section \ref{sec: h4}. In this open-loop scenario, the agent learns about the unmet tasks only after the initial plan $\pi^{*}_{i}$ fails to achieve the goal states, meaning it can only replan to complete the remaining tasks following the initial plan's failure, as opposed to the immediate adaptation observed in our framework. This approach is inefficient because, even if the agent is given ample time, it may complete the goal states, but at the expense of a substantially higher execution cost.

\color{black}
%%%%%%%%%%%%%%%%%%%%%%%%%%%%%%%%%%%%%%%%%%%%%%%%%%%%%%%%%%%%%%%%%%%%%%%%%%%%%%%%%%%%%%%%% 
\subsection{Experimental Results}
\label{sec:expres-results}
%%%%%%%%%%%%%%%%%%%%%%%%%%% EXPERIMENTAL RESULTS %%%%%%%%%%%%%%%%%%%%%%%%%%%%%%%%%%%%%%%%
%We now describe the results of our experimental analysis.

\noindent
\textbf{Evaluating H1.} As stated in Section~\ref{sec:expres-setup}, we evaluated the ability of three different LLMs to anticipate upcoming tasks in two different households, compared these predictions with those of a set of humans, and used Equations~\ref{eqn:mean-overlap}-\ref{eqn:overlap} as the performance measures. We also considered the two prompting mechanisms described in Section~\ref{sec:framework-llm}. The results are documented in Table~\ref{tab:llm_overlap_gt}. Recall that \textit{household-1} was characterized by an orderly task execution whereas \textit{household-2} prioritized immediate needs. The results in Table~\ref{tab:llm_overlap_gt} indicate that there was a greater degree of overlap between the predictions of the LLMs and those of humans in household-1. In the case of household-2, there was greater variability in the predictions provided by the LLMs and in those provided by the human subjects; as a result, the degree of overlap between humans and LLMs was lower.

\vspace{-1em}
Among the prompting methods, better performance, i.e., a higher degree of overlap between the predictions of the LLMs and the humans, was obtained with the chain of thought prompting compared with the few shot prompting method. In fact, the mean overlap value was as high as $86\%$ for the \textit{Claude-3} LLM, and it correctly anticipated at least three out of four tasks with the chain of thought prompting (compared with the ground truth). In addition, in the absence of this prompting method, which includes more contextual information, the LLMs often went into a loop of hallucination. Overall, these results support hypothesis \textbf{H1}.

\begin{table}[tb]
    \centering
    \captionsetup{font=scriptsize}
        \begin{adjustbox}{width=\columnwidth,center}
            \begin{tabular}{|c|l|c|c|c|c|c|c|}
                \hline
                & \multirow{2}{1.9cm}{LLM Models $\rightarrow$}  & \multicolumn{2}{c|}{Claude} & \multicolumn{2}{c|}{GPT-4} & \multicolumn{2}{c|}{Gemini} \\[0.1em]
                \cline{3-8}
                & & few-shot & CoT & few-shot & CoT & few-shot & CoT \\ [0.2em]
                \hline
                \multirow{3}{1.5cm}{Household 1}& Overlap & 0.72 & \textbf{0.86} & 0.75 & 0.82 & 0.59 & 0.77 \\ [0.2em]
                \cline{2-8}
                & $\geq50\%$ overlap & 0.99 & \textbf{1} & \textbf{1} & \textbf{1} & 0.92 & \textbf{1} \\ [0.2em]
                \cline{2-8}
                & $\geq75\%$ overlap & 0.77 & \textbf{1} & 0.8 & 0.96 & 0.46 & 0.78 \\ [0.2em]
                \hline
                \multirow{3}{1.5cm}{Household 2}& Overlap & 0.70 & 0.71 & 0.68 & \textbf{0.72} & 0.57 & 0.6 \\ [0.2em]
                \cline{2-8}
                & $\geq50\%$ overlap & 0.94 & 0.98 & 0.96 & \textbf{1} & 0.9 & 0.93 \\ [0.2em]\cline{2-8}
                & $\geq75\%$ overlap & 0.69 & \textbf{0.79} & 0.68 & 0.76 & 0.32 & 0.42 \\ [0.2em]
                \hline
            \end{tabular}
        \end{adjustbox}
    \caption{Evaluating LLM-based task anticipation for two separate households based on few-shot prompting and chain-of-thought reasoning. Results support \textbf{H1}.}
    \label{tab:llm_overlap_gt}
\end{table}

\begin{figure}[tb]
  \centering
  \captionsetup{font=scriptsize}
  \includegraphics[width=\columnwidth]{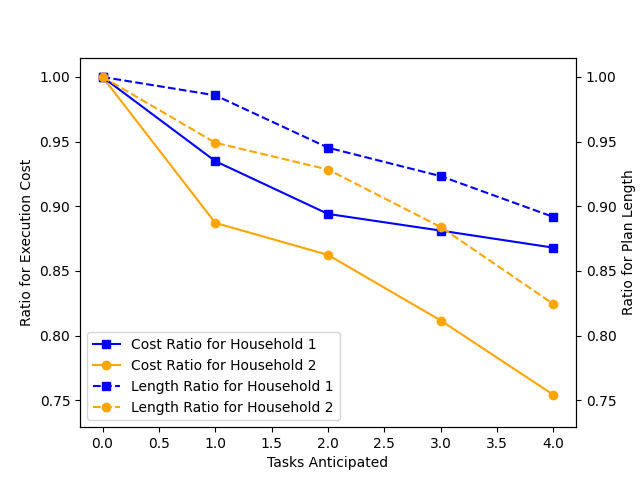}
  \caption{Evaluating \textbf{H2}. Values of \textit{execution cost} and \textit{plan length} with different levels of anticipation computed as a ratio over values computed for no anticipation; paired trials conducted for different scenarios in two different households. The combination of task anticipation and action planning improved performance.}
  \label{fig:h2_plot_combined}
  \vspace{-1em}
\end{figure}

\noindent
\textbf{Evaluating H2.}
We compared the performance of DaTAPlan (i.e., our framework) with that of a framework that used a classical planner to compute the sequence of actions for one high-level task at a time. As with \textbf{H1}, we considered the five scenarios in each of two households. In each scenario, the number of anticipated tasks was varied from 0-4, with 0 corresponding to no anticipation. The planner was provided an initial search time limit that was then increased in fixed increments. We conducted paired trials, i.e., for any particular scenario in a specific household, we would pick a particular initial state and run trials with and without anticipation. Recall that the performance measures for this experiment were \textit{execution cost} and \textit{plan length}. Since the values of these measures can change drastically depending on the initial conditions, we computed the ratio of the value (of a performance measure) for a given level of anticipation (ranges from 1-4 tasks being anticipated) with the value for no anticipation. The results are summarized in Figure~\ref{fig:h2_plot_combined}; each point in the plots is the average of the ratios computed for five scenarios in a particular household.

\vspace{-1em}
The results in Figure~\ref{fig:h2_plot_combined} clearly indicated a decrease in the planning time and execution cost as the number of anticipated tasks increased. Overall, there was a drop in total execution cost of $\approx 12.5\%$ and $\approx 25\%$ in household-1 and household-2 respectively as the number of anticipated tasks increases from one to four. Similarly, we observed a decrease of approximately $10\%$ and $17.5\%$ in plan length in Household 1 and 2 respectively. These plots support hypothesis \textbf{H2} and demonstrate the advantages of integrating LLM-based high-level task anticipation with planning (a sequence of low-level actions) for achieving the tasks.

\begin{table*}[tb]
\centering
\captionsetup{font=scriptsize}
\setlength{\belowcaptionskip}{-7pt}
\begin{tabular}{| >{\centering\arraybackslash} m{1.0cm}| 
>{\centering\arraybackslash} m{1.5cm}| >{\centering\arraybackslash} m{1.0cm}| >{\centering\arraybackslash} m{1.0cm}|  >{\centering\arraybackslash} m{1.0cm}| >{\centering\arraybackslash} m{1.0cm}| >{\centering\arraybackslash} m{1.0cm}| >{\centering\arraybackslash} m{1.0cm}| >{\centering\arraybackslash} m{1.0cm}| >{\centering\arraybackslash} m{1.0cm}|}
\hline
\\[-1em]
  & \multicolumn{9}{c|}{ROBOT} \\
\cline{1-10}
\\[-1em]
 \multirow{6}{1.0cm}{HUMAN}&  Init $\rightarrow$ & \multicolumn{2}{c|}{Livingroom} & \multicolumn{2}{c|}{Storeroom} & \multicolumn{2}{c|}{Bathroom} & \multicolumn{2}{c|}{Kitchen}  \\
\cline{2-10}
\\[-1em]
 &  Init $\downarrow$ & $\zeta_{HFAS}$ & $\zeta_{AFHS}$ & $\zeta_{HFAS}$ & $\zeta_{AFHS}$ & $\zeta_{HFAS}$ & $\zeta_{AFHS}$ & $\zeta_{HFAS}$ & $\zeta_{AFHS}$ \\
\cline{2-10}
\\[-1em]
 & Livingroom & 0.913 & 0.778 & 0.883 & \textbf{0.747} & 0.913 & 0.788 & 0.895 & 0.757\\
% \\[-1em]
\cline{2-10}
\\[-1em]
&  Storeroom & 0.969 & 0.873 & 0.959 & 0.863 & 0.98 & 0.843 & 0.929 & 0.853 \\
\cline{2-10}
\\[-1em]
&  Bathroom & 0.968 & 0.838 & 0.926 & 0.794 & 0.949 & 0.828 & 0.911 & 0.833 \\
\cline{2-10}
\\[-1em]
&  Kitchen & 0.943 & 0.846 & 0.927 & 0.81 & 0.925 & 0.826 & 0.924 & 0.856 \\
% & \multirow{4}{1.5cm}{seq-sat-fdss-2018 (90 sec)} & Livingroom & 1.143 & 1.171 & 1.09 & \textbf{0.972} & 1.058 & 1.134 & 1.063 & 1.009 \\
% \cline{3-10}
% \\[-1em]
% &  & Storeroom & 1.155 & 1.066 & 1.238 & 1.135 & 1.149 & 1.149 & 1.167 & 1.185 \\
% \cline{3-10}
% \\[-1em]
% &  & Bathroom & 1.196 & 1.108 & 1.162 & 1.053 & 1.203 & 1.126 & 1.23 & 1.163 \\
% \cline{3-10}
% \\[-1em]
% &  & Kitchen & 1.202 & 1.105 & 1.169 & 1.025 & 1.139 & 1.045 & 1.176 & 1.112 \\
% \cline{2-10}
% \\[-1em]
% & \multirow{4}{1.5cm}{seq-sat-fd-autotune-1 (90 sec)} & Livingroom & 1.039 & \textbf{0.92} & 1.016 & \textbf{0.909} & 0.906 & \textbf{0.9} & 0.806 & \textbf{0.914} \\
% \cline{3-10}
% \\[-1em]
% &  & Storeroom & 1.024 & \textbf{0.987} & 1.089 & 1.005 & 1.052 & \textbf{0.997} & 1.055 & 1.021 \\
% \cline{3-10}
% \\[-1em]
% &  & Bathroom & 1.013 & \textbf{0.948} & 1.043 & \textbf{0.949} & 1.067 & \textbf{0.946} & 1.025 & \textbf{0.936} \\
% \cline{3-10}
% \\[-1em]
% &  & Kitchen & 1.038 & \textbf{0.982} & 1.05 & \textbf{0.981} & 1.043 & \textbf{0.968} & 1.091 & 1.026 \\
\hline
\end{tabular}
\caption{Evaluating \textbf{H3}. Computed $\zeta$ as the ratio of execution time with collaboration over the execution time without collaboration. Each value in table is average over 50 trials (with two of more tasks in the goal state) for each of 16 possible initial locations of the agent and the human, and for each of two settings: HFAS and AFHS. Results indicate a clear benefit of human-agent collaboration in terms of reduction in execution time, thus supporting \textbf{H3}.}
\label{tab:h3_collab_table}
\end{table*}

\noindent
\textbf{Evaluating H3.} 
We conducted paired trials with and without collaboration, and measured the total time taken by both actors (agent, human) to achieve the corresponding goal states. Specifically, we created 16 different initial conditions by considering different combinations of initial state of the robot and the human, e.g., agent initially in the \textit{Kitchen} with human in the \textit{Storeroom}. We also considered two settings: (i) \textit{HFAS (Human-Fast-Agent-Slow)}, in which the human was faster than the robot and had a lower movement cost; and (ii) \textit{AFHS (Agent-Fast-Human-Slow)}, in which the agent was faster and had a lower movement cost. Then, for each initial condition and each setting, we ran 50 trials, each with two tasks in the goal state. In each trial, we computed $\zeta$ as the ratio of the total execution time with collaboration over the value of the measure without collaboration; $\zeta < 1$ denotes effective collaboration. We then computed the average of these ratios for a particular initial state, and report the average of these ratios in  Table~\ref{tab:h3_collab_table}.

\vspace{-1em}
Results indicate that we achieved $\zeta$ < 1 for all initial conditions and all settings, clearly demonstrating the benefits of collaboration between the agent and the human in terms of the reduction in execution time; the maximum reduction in time (or equivalently the cost) was $25.3\%$, which was obtained in the AFHS setting. Overall, these results clearly support hypothesis \textbf{H3}.

\begin{figure}[tb]
    \centering
    \captionsetup{font=scriptsize}
    \begin{tabular}{|c|c|}
        \hline
        & \textbf{Overlap} \\ \hline
        Gemini & 0.59 \\ \hline
        Claude & 0.98 \\ \hline
        GPT-4 & 0.91 \\ \hline
    \end{tabular}
\captionof{table}{Evaluating \textbf{H4}. Examined re-prompting of different LLMs during task execution in response to an unexpected change in human preference that makes the current set of anticipated tasks irrelevant. We measured the overlap between new set of anticipated tasks predicted by LLMs and the new predictions provided by human subjects; results support \textbf{H4}.}
\label{tab:llm_reprompt}
\vspace{-1.5em}
\end{figure}
%\end{wrapfigure}

\noindent
\textbf{Evaluating H4.}
% \label{sec: h4}
We first explored the ability of the agent to adapt to unexpected changes in human preferences, i.e., in situations that required the agent to generate a new set of anticipated tasks (from the LLM) before computing and executing the corresponding plan. In these experiments, the interrupt (due to change in human preference) occurs as the agent and the human are executing the actions.
%In Hypothesis \textbf{H1}, we evaluated the effectiveness of LLMs in anticipating future tasks based on the pattern of task execution in a house. In this experiment, we measure the LLMs in cases when the task execution differs from the usual pattern, and hence the anticipation is expected to go wrong. 
We conducted experiments similar to those for \textbf{H1}, measuring the overlap between the new set of anticipated tasks predicted by the LLMs and the tasks predicted by human subjects; we conducted these experiments over 25 repetitions (each) of five different scenarios (with chain of thought promoting). The corresponding results, summarized in Table~\ref{tab:llm_reprompt}, indicated support for hypothesis \textbf{H4}.

\vspace{-0.5em}
Next, we explored the ability to adapt to unexpected changes in the outcomes of human actions; as stated in Section~\ref{sec:framework-tamp}, we randomly introduced errors in the outcome of the \textit{pick up} actions executed by the human. We considered combinations of 16 different goals, four each with 1-4 high-level tasks, and 1-3 instances of errors (in human pickup actions). 
The goal states contained tasks ranging from simple ones such as picking up clothes from the closet and placing them on the ironing board, to complex tasks such as \textit{cleaning} and \textit{slicing} multiple fruits before placing them in a \textit{bowl} to \textit{prepare salad}. \textit{Our adaptation approach results in the goals being accomplished in all trials despite the errors in human actions} by replanning from a revised initial state, but there is an increase in execution (and planning) time. We thus conducted paired trials and computed (in each trial) the fraction of the high-level tasks (in the goal) achieved in absence of our adaptation approach in the same amount of time taken by our adaptation approach. The corresponding results are summarized in Table~\ref{tab:multiple_noise_table}, with the last value in each row representing the average performance over different levels of errors for a goals with a specific number of high-level tasks. There was a reduction in performance in the absence of our adaptation approach, and this reduction was (understandably) more pronounced with the increase in the number of instances of errors in the outcomes of the human actions. These results support \textbf{H4}.

% We have used total 16 different goal states for our experiments. The goal states used here contain tasks ranging from very simple, like {\small \texttt{(prepared\_clothes office\_clothes ironing\_board livingroom)}}, which involves picking up clothes from the closet and placing them on the ironing board, to very complex tasks like {\small \texttt{(salad\_prepared bowl\_1 sliced\_apple sliced\_avocado sliced\_banana)}}, which involve \textit{cleaning} and \textit{slicing} fruits before placing them in a \textit{bowl}.

\begin{table}[tb]
    \centering
    \captionsetup{font=scriptsize}
        \begin{adjustbox}{width=\columnwidth,center}
            \begin{tabular}{|c|c|c|c|c|}
                \hline
                 & \multicolumn{3}{c|}{ No. of erroneous outcomes} & \multirow{2}{1.0cm}{Success Rate \%} \\[0.25em]
                \cline{2-4}
                % \multirow{2}{1.0cm}{No. of Tasks}& 1 & 2 & 3 \\ [0.25em]
                % \cline{2-4}
                No. of Tasks& 1 & 2 & 3 & \\ [0.25em]              
                  % & Fraction of task completed & Fraction of task completed  & Fraction of task completed  \\ [0.25em]
                \hline
                 1 & 0 $\pm$ 0 & 0 $\pm$ 0 & 0 $\pm$ 0 & 0.0 $\pm$ 0.0 \\ [0.25em]
                \hline
                 2 & 0.5 $\pm$ 0 & 0.25 $\pm$ 0.25 & 0 $\pm$ 0 & 25.0 $\pm$ 8.33  \\ [0.25em]
                \hline
                 3 & 0.49 $\pm$ 0.16 & 0.33 $\pm$ 0.23 & 0.25 $\pm$ 0.14 & 36.11 $\pm$ 9.21 \\ [0.25em]
                \hline
                 4 & 0.62 $\pm$ 0.12 & 0.56 $\pm$ 0.11 & 0.43 $\pm$ 0.11 & 54.17 $\pm$ 9.32\\ [0.25em]
                \hline
            \end{tabular}
        \end{adjustbox}

    \caption{Evaluating \textbf{H4}. Considered combinations of goal states with 1-4 high-level tasks with 1-3 instances of erroneous human (pick up) actions. Computed mean and standard deviation of the fraction of tasks completed without adaptation in the same amount of time taken by the adaptation approach (in DaTAPlan) to achieve the goal. Results support \textbf{H4}.}
    \label{tab:multiple_noise_table}
    \vspace{-1em}
\end{table}

% Figure \ref{fig: No_Adaptation_2} illustrates a plot with 16 different goal states divided into four categories based on the number of tasks they contain. The first category comprises goal states with a single task, the second with two tasks, and the third and fourth with three and four tasks, respectively. The y-axis indicates the percentage of tasks completed in scenarios where the system is non-adaptive to noise. For goal states with a single task, there are only two possible outcomes: 0\% and 100\%. Achieving the full goal state is not possible due to the presence of noise.

% \begin{figure}[h!]
%   \centering
%   \captionsetup{font=scriptsize}
%   \includegraphics[width=1
% \linewidth]{figures/output_4.png}
%   \caption{
% The plot displays the percentage of tasks completed in a non-adaptive scenario. It measures the proportion of tasks accomplished in the non-adaptive scenario when the allotted time is equivalent to that in the adaptive scenario.}
%   \label{fig: No_Adaptation_2}
% \end{figure}

\begin{figure}[tb]
  \centering
  \captionsetup{font=scriptsize}
  \includegraphics[width=\columnwidth]{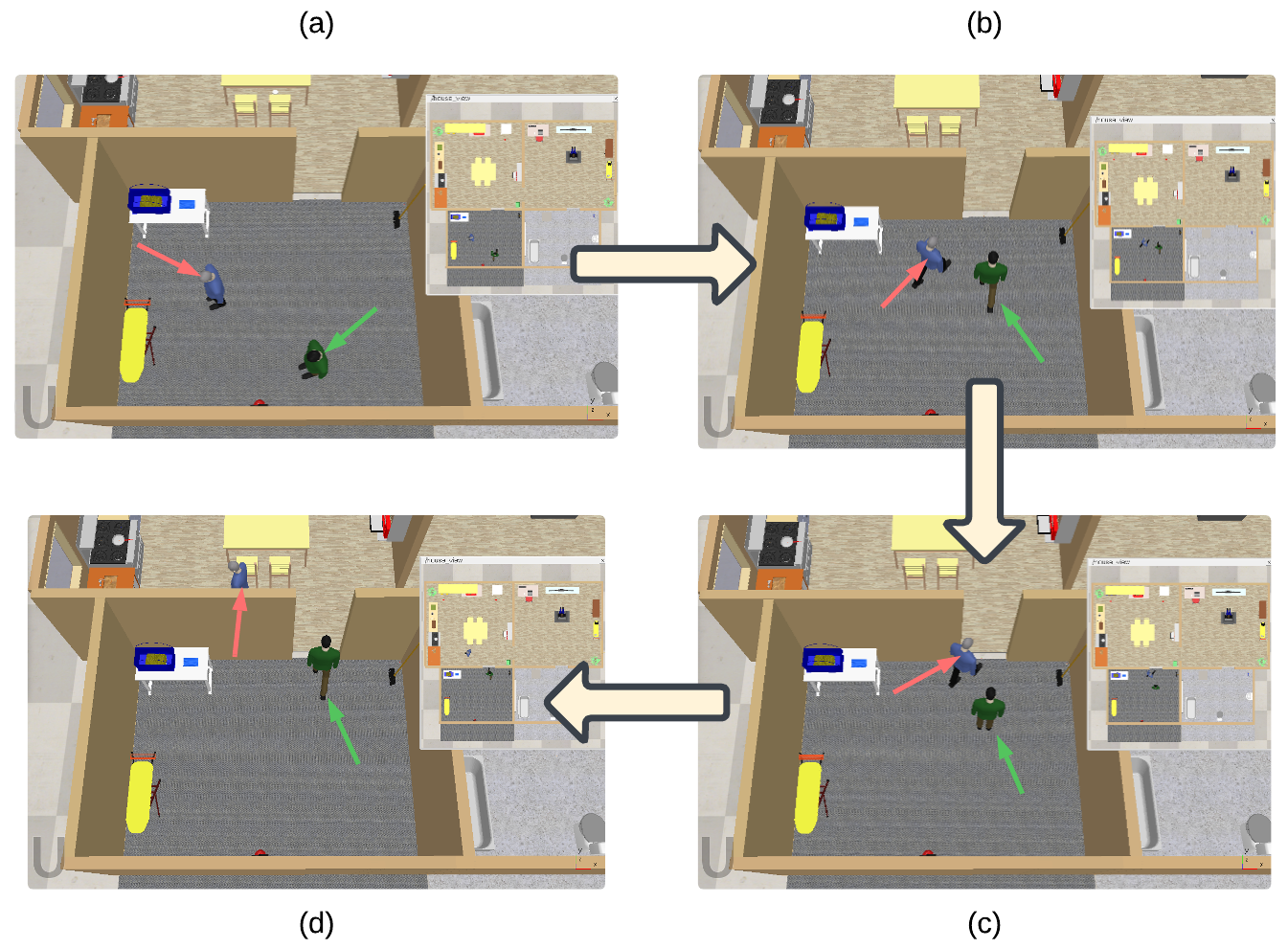}
  \caption{Collision avoidance between human (orange arrow) and the agent (green arrow) in CoppeliaSim: (a) initial position of human and agent; (b) both move in a direction that could result in a collision in the near future; (c) agent stops and waits for the human to pass; (d) agent continues once the collision situation is addressed.}
  \label{fig:collision_avoidance}
  \vspace{-1em}
\end{figure}

\vspace{-1em}
Finally, we evaluated the ability of our framework to detect and address instances of possible collisions between the agent and the human. Any such potential collisions were successfully avoided in all trials and we show a qualitative example in Figure~\ref{fig:collision_avoidance}. Our implementation successfully tracks the trajectory of both actors. When a potential collision is detected, the agent stops and allows the human to pass first, and proceeds with its trajectory once it is safe to do so again.

% Our system has accounted for a local response to possible collisions between the human and the agent in the environment which can be seen in Fig. \ref{fig:collision_avoidance}. The trajectory of both the actors was tracked and a feedback response to stop was generated at the agent level. Once the obstruction in the agent's trajectory is removed, the agent restarts to follow it's generated trajectory.

% Human Fast Agent Slow:
% Autotune1: [1.067, 1.025, 1.013, 1.043, 1.043, 1.091, 1.038, 1.05, 1.005, 1.024, 1.039, 1.016, 1.052, 1.055, 1.024, 1.089]
% FDSS-2018: [1.203, 1.23, 1.196, 1.162, 1.139, 1.176, 1.202, 1.169, 1.058, 1.063, 1.143, 1.09, 1.149, 1.167, 1.155, 1.238] Lama: [0.949, 0.911, 0.968, 0.926, 0.925, 0.924, 0.943, 0.927, 0.913, 0.895, 0.913, 0.883, 0.98, 0.929, 0.969, 0.959]

% Agent Fast Human Slow:
% Autotune-1: [0.946, 0.936, 0.948, 0.949, 0.968, 1.026, 0.982, 0.981, 0.9, 0.914, 0.92, 0.909, 0.997, 1.021, 0.987, 1.005]
% FDSS-2018: [1.126, 1.163, 1.108, 1.053, 1.045, 1.112, 1.105, 1.025, 1.134, 1.009, 1.171, 0.972, 1.149, 1.185, 1.066, 1.135]
% Lama: [0.828, 0.833, 0.838, 0.794, 0.826, 0.856, 0.846, 0.81, 0.788, 0.757, 0.778, 0.747, 0.843, 0.853, 0.873, 0.863]

\section{Conclusions and Future Works}
This paper described DaTAPlan, a framework that combines data-driven task anticipation with knowledge-driven planning for reliable and efficient human-agent collaboration. This builds on our recent work that combined LLM-based task anticipation with classical planning for a single agent in a domain with no other actors~\cite{anticipate_act}. Here, the LLM-based anticipation is extended to accommodate different user patterns. Also, the anticipated high-level tasks form the goal for a classical planning system that generates a sequence of finer-granularity actions, with both the agent and the human executing actions to collaboratively achieve each task in the goal. In addition, an approach enables adaptation to unexpected changes in the action outcomes and preferences of the human. We experimentally evaluated these capabilities and demonstrated substantial improvement in performance compared with a framework without collaboration or without the adaptation approach.

% We show that language models are able to retrieve user-specific behaviours given a few observations, and anticipating future tasks leads to a cost reduction of upto $25\%$. We explore the ability of different language models to adapt to the patterns of specific individuals, and observe that Claude 3 outperforms other language models in matching human expectations. 
% We further demonstrate human-robot collaboration during execution of lower-granularity actions to jointly achieve anticipated goals in the house. 
% We observe a reduction in execution time of upto $25.3\%$ when the human and the robot collaborate together, and demonstrate that our framework can account for deviations from anticipated tasks, and deviation of the human from expected actions during execution. 
% We show the deviation of human from the expected plan only while picking up objects in the home. 
\vspace{-1em}
The paper opens up multiple directions of further research. First, we currently do not model or support any active communication between the agent and the human; this constraint can be relaxed to consider communication as an action. Second, we consider the domain descriptions (including action costs) to be complete and accurate; future work can explore reasoning with incomplete descriptions and the incremental revision of existing descriptions. Third, action execution has limited uncertainty in the current implementation and there is full observability; these assumptions can be relaxed in the future. Overall, the long-term objective is to enable physical robots to provide reliable and efficient assistance to humans in complex domains.

% The future work of this paper would involve considering divergence of the human for other actions, and incorporating probabilistic planning. 
% We intend to use learning methods to update action costs on the fly based on collaboration with the human.
\balance

\bibliographystyle{IEEEtran}
\bibliography{IEEEabrv,references}
\end{document}